\documentclass[conference]{IEEEtran}
\IEEEoverridecommandlockouts
\usepackage{cite}
\usepackage{amsmath,amssymb,amsfonts}
\usepackage{algorithmic}
\usepackage{graphicx}
\usepackage{textcomp}
\usepackage{xcolor}
\usepackage{listings}

\colorlet{punct}{red!60!black}
\definecolor{background}{HTML}{EEEEEE}
\definecolor{delim}{RGB}{20,105,176}
\colorlet{numb}{magenta!60!black}

\lstdefinelanguage{json}{
    basicstyle=\normalfont\ttfamily,
    numbers=left,
    numberstyle=\scriptsize,
    stepnumber=1,
    numbersep=8pt,
    showstringspaces=false,
    breaklines=true,
    frame=single,
    backgroundcolor=\color{background},
    literate=
     *{0}{{{\color{numb}0}}}{1}
      {1}{{{\color{numb}1}}}{1}
      {2}{{{\color{numb}2}}}{1}
      {3}{{{\color{numb}3}}}{1}
      {4}{{{\color{numb}4}}}{1}
      {5}{{{\color{numb}5}}}{1}
      {6}{{{\color{numb}6}}}{1}
      {7}{{{\color{numb}7}}}{1}
      {8}{{{\color{numb}8}}}{1}
      {9}{{{\color{numb}9}}}{1}
      {:}{{{\color{punct}{:}}}}{1}
      {,}{{{\color{punct}{,}}}}{1}
      {\{}{{{\color{delim}{\{}}}}{1}
      {\}}{{{\color{delim}{\}}}}}{1}
      {[}{{{\color{delim}{[}}}}{1}
      {]}{{{\color{delim}{]}}}}{1},
}

\usepackage[most]{tcolorbox}
\tcbuselibrary{theorems}

\usepackage[font=footnotesize,skip=6pt]{caption}

\usepackage[most]{tcolorbox}
\tcbuselibrary{theorems}
\usepackage{caption}

\newtcbtheorem{example}{Example}%
  {enhanced,
   colback=white,
   colframe=blue!30!black,
   coltitle=white,
   attach boxed title to top left={xshift=3mm,yshift*=-3mm},
   boxed title style={colback=blue!30!black}
  }{ex}

\def\BibTeX{{\rm B\kern-.05em{\sc i\kern-.025em b}\kern-.08em
    T\kern-.1667em\lower.7ex\hbox{E}\kern-.125emX}}
\begin{document}

\title{Quadrupped-Legged Robot Movement Plan Generation using Large Language Model\\
\thanks{This research was financially supported by the Final Project Assistance Grant (Bantuan Tugas Akhir Mahasiswa) funded by Institut Teknologi Sepuluh Nopember (ITS) under the 2025 ITS Internal Research Grant Scheme.  }
}

\author{%
Muhtadin$^{1}$$^{,}$$^{2}$,
Vincentius Gusti Putu A. B. M.$^{2}$,
Ahmad Zaini$^{2}$,
Mauridhi Hery Purnomo$^{1}$$^{,}$$^{2}$,
I Ketut Eddy Purnama$^{1}$, \\
Chastine Fatichah$^{3}$ \\
\small $^{1}$Department of Electrical Engineering, Institut Teknologi Sepuluh Nopember, Surabaya, Indonesia \\
\small $^{2}$Department of Computer Engineering, Institut Teknologi Sepuluh Nopember, Surabaya, Indonesia \\
\small $^{3}$Department of Informatics, Institut Teknologi Sepuluh Nopember, Surabaya, Indonesia \\
Email: muhtadin@its.ac.id,  vincentiusgt23@gmail.com, zaini@its.ac.id, hery@ee.its.ac.id, ketut@te.its.ac.id, chastine@if.its.ac.id
}

\maketitle

\begin{abstract}
Traditional control interfaces for quadruped robots often impose a high barrier to entry, requiring specialized technical knowledge for effective operation. To address this, this paper presents a novel control framework that integrates Large Language Models (LLMs) to enable intuitive, natural language-based navigation. We propose a distributed architecture where high-level instruction processing is offloaded to an external server to overcome the onboard computational constraints of the DeepRobotics Jueying Lite 3 platform. The system grounds LLM-generated plans into executable ROS navigation commands using real-time sensor fusion (LiDAR, IMU, and Odometry). Experimental validation was conducted in a structured indoor environment across four distinct scenarios, ranging from single-room tasks to complex cross-zone navigation. The results demonstrate the system's robustness, achieving an aggregate success rate of over 90\% across all scenarios, validating the feasibility of offloaded LLM-based planning for autonomous quadruped deployment in real-world settings.
\end{abstract}


\section{Introduction}
Quadruped robots have emerged as versatile platforms capable of navigating complex terrains \cite{Zhou_Adaptive, Han_Lifelike, Mastalli_Motion}. However, their widespread deployment is often hindered by conventional control interfaces, which are typically unintuitive and require specialized technical expertise, creating a significant barrier for non-expert users \cite{Nwankwo_conversation,Wilde_Improving}. Large Language Models (LLMs) offer a transformative solution to this challenge by enabling intuitive, natural language interaction \cite{Cai_Low-code}. Beyond simple communication, LLMs function as effective high-level planners capable of "grounding" abstract instructions into actionable robotic sequences, a capability famously demonstrated by the "SayCan" framework \cite{Ahn_Do}. This bridge between language and action enables robots to interpret intricate tasks, effectively translating human intent into precise robotic movement.

Recent research has successfully applied LLM-based control to quadruped robots across various domains. For instance, Ginting et al. \cite{Ginting_comply} utilized LLMs for indoor inspection missions, while Balaji et al. \cite{Balaji_Language} demonstrated their application for object search in agricultural environments. Similarly, Anwar et al. \cite{Anwar_Remembr} employed LLMs to guide visitors through complex spaces using landmark-based navigation. Despite these advancements, a critical architectural challenge remains: the high computational demand of LLMs often exceeds the resources of standard mobile robot hardware \cite{Waga_survey}, \cite{Lin_Advances}. Existing solutions typically rely on high-performance onboard GPUs or static server installations, which are not feasible for lightweight or cost-effective mobile platforms.

To address this limitation, this paper proposes a distributed control architecture for the DeepRobotics Jueying Lite 3 robot. Unlike fully onboard solutions, our system offloads the computationally intensive LLM inference to an external server while maintaining robust, real-time local navigation via ROS. This architecture ensures that the robot can process and execute complex, natural language-driven navigation tasks in structured indoor environments without compromising its limited onboard computational capabilities. The effectiveness of this approach is validated through real-world experiments, demonstrating high success rates in multi-room and cross-zone navigation tasks.

Building upon our prior research on service robots—which established capabilities in mapping \cite{Muhtadin_Autonomous}, object following \cite{Muhtadin_Implementation}, and lost item retrieval for elderly care \cite{Muhtadin_Robot}—this study serves as a complementary step toward a holistic service architecture. By integrating LLM-driven planning, we aim to bridge the gap between functional autonomy and intuitive human-robot interaction.

% The IEEEtran class file is used to format your paper and style the text. All margins, 

\begin{figure*}[t]
\centerline{\includegraphics[scale=0.40]{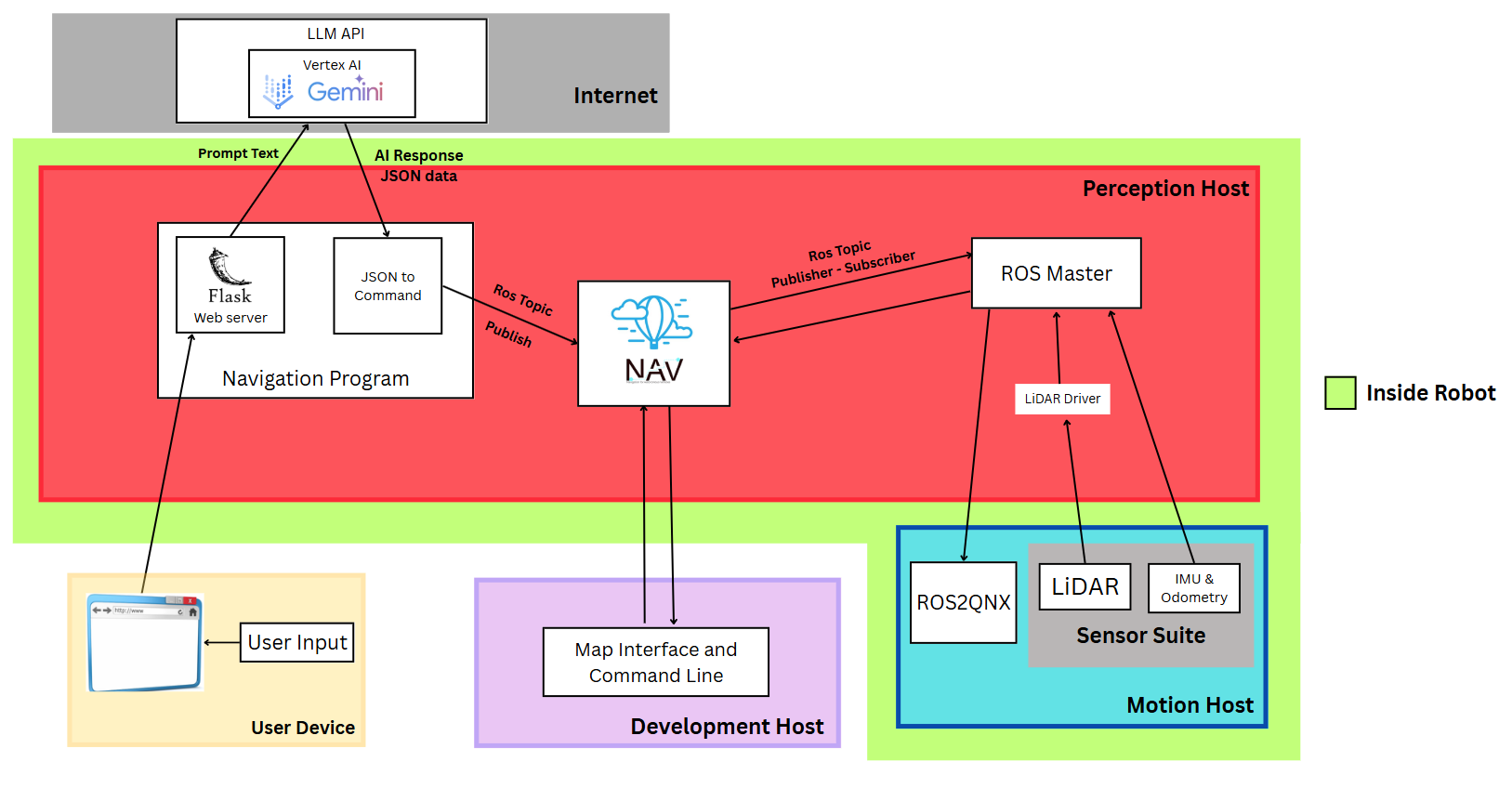}}
\caption{Movement Plan generation system architecture}
\label{system_architecture}
\end{figure*}

\section{System Design and Configuration}
Our control system was configured and implemented on the DeepRobotic Lite 3 robot, which has two main computers called the motion host and the perception host. The motion host is responsible as the main connection to the robot motion actuator and sensor fleet, such as the LiDAR, IMU sensor, and odometry sensor. On the other hand, the perception host is mainly responsible for handling complex instructions with more capable computing power, consisting of Nvidia Jetson NX Xavier. All the sensor fusion and processing used for localization and planning are running inside the perception host. 

\subsection{Distributed Hardware Architecture}\label{AA}
Fig. \ref{system_architecture} describes the architecture and configuration of the control system designed for the quadruped-legged robot that utilizes the Large Language Model (LLM). The system is divided into five key components: the user device, development host, perception host, motion host, and internet access point. These components work together to interpret human natural language instructions, generate a movement plan, and control the robot autonomously in a structured environment.

The control framework involves a distributed architecture wherein the perception and motion hosts reside directly on the robot platform, while the development host is responsible for handling LLM requests. The motion and development hosts communicate via a LAN cable, and the development host connects to a local router within the robot to act as a Wi-Fi access point. This allows external user devices (smartphones or computers) on the same network, to access the Flask-based web server hosted by the robot.

The sequence of operations begins with the initialization of the robot's sensors—particularly the LiDAR and IMU—which are activated via remote desktop access to the perception host. Once the sensor data is available to the ROS Master, the user launches the ROS navigation stack and localizes the robot using RViz by placing it on both the 2D and 3D maps.

Following successful localization, the navigation program begins publishing movement commands to the navigation stack via ROS topics. Concurrently, a Flask web server is activated to provide a user-friendly interface for natural language input. Users can submit instructions in Indonesian Language, which are processed through the LLM to generate a structured JSON-based movement plan. These commands are subsequently parsed and executed by the robot's navigation controller.

\subsection{Mapping and Navigation}
\begin{center}
\begin{figure*}[t]
\includegraphics[scale=0.70, trim=0 12cm 5cm 0]{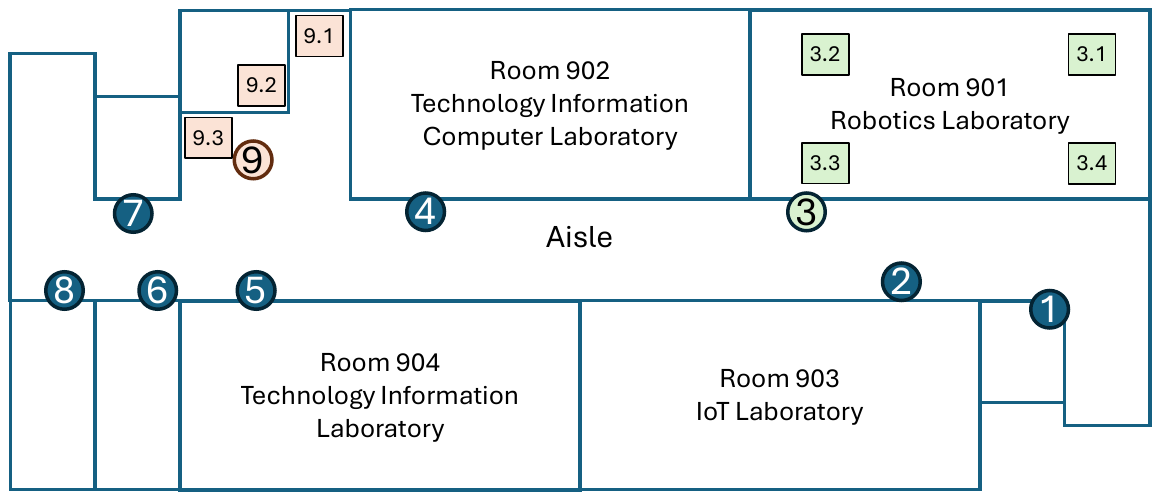}
\caption{Point of interest in the main hall}
\label{fig:mapseluruhnya}
\end{figure*}
\end{center}

Effective path planning requires precise environmental awareness. We utilize HDL-Localization, a 3D LiDAR-based SLAM technique, to construct a high-fidelity map of the structured indoor environment (Tower 2 Building, ITS Campus). Within this map, we define semantic waypoints representing key points of interest \(POIs\) such as laboratories, pantries, and elevators, described in Fig. \ref{fig:mapseluruhnya} and Table \ref{tabel:semanticpoint}. Each waypoint \(W_i\) is associated with a global coordinate \(x,y,z\) in the map frame, allowing the navigation stack to execute point-to-point movement using global planning algorithms.

\begin{table}[t]
\caption{Semantic Waypoints and Interior Zones}
\label{tabel:semanticpoint}
\begin{center}
\begin{tabular}{|c|l|p{3.5 cm}|}
\hline
\textbf{ID} & \textbf{Location Name} & \textbf{Specific Indoor Areas \newline (Points of Interest)} \\
\hline
\textbf{1} & Elevator  & Elevator Door / Waiting Area \\
\hline
\textbf{2} & IoT Lab (Room 903) & Room 903 Door \\
\hline
\textbf{3} & Robotics Lab (Room 901) & \textbf{3.1} Robot Home Pos. \newline \textbf{3.2} Assembly Table \newline \textbf{3.3} Lab Shelf \newline \textbf{3.4} Soldering Table \\
\hline
\textbf{4} & Computer Lab (Room 902) & Room 902 Door \\
\hline
\textbf{5} & IT Lab (Room 904) & Room 904 Door \\
\hline
\textbf{6} & Men's Restroom & Restroom Entrance \\
\hline
\textbf{7} & Elevator  & Elevator Door / Waiting Area \\
\hline
\textbf{8} & Women's Restroom & Restroom Entrance \\
\hline
\textbf{9} & Pantry Area & \textbf{9.1} Security Room \newline \textbf{9.2} Pantry Kitchen \newline \textbf{9.3} Pantry Shelf \\
\hline
\end{tabular}
\label{tab:waypoints}
\end{center}
\end{table}

\subsection{LLM Prompt Design and Integration}
The core of our intuitive interface is the translation of natural language into structured robotic commands. We employ a carefully engineered system prompt that instructs the LLM (Vertex AI Gemini) to function as a motion planner. The prompt enforces a strict JSON output format containing an array of "actions," where each action consists of a command (e.g., goto, wait) and relevant parameters (e.g., waypoint: "pantry"). To ensure reliability, the prompt includes:
\begin{enumerate}
    \item Action Primitives: A defined list of valid robot behaviors (navigation, exploration, halting).

    \item Contextual Constraints: Rules preventing the generation of hallucinated or unsafe waypoints. 
    \item Few-Shot Examples: Input-output pairs guiding the model to parse complex, multi-step instructions into sequential JSON objects. The generated JSON is parsed by the Development Host and published to the ROS move\_base topic for execution.

\end{enumerate}
    
\subsection{Web Interface and LLM API Integration}
\begin{figure}[t]
\centering
\includegraphics[scale=0.18]{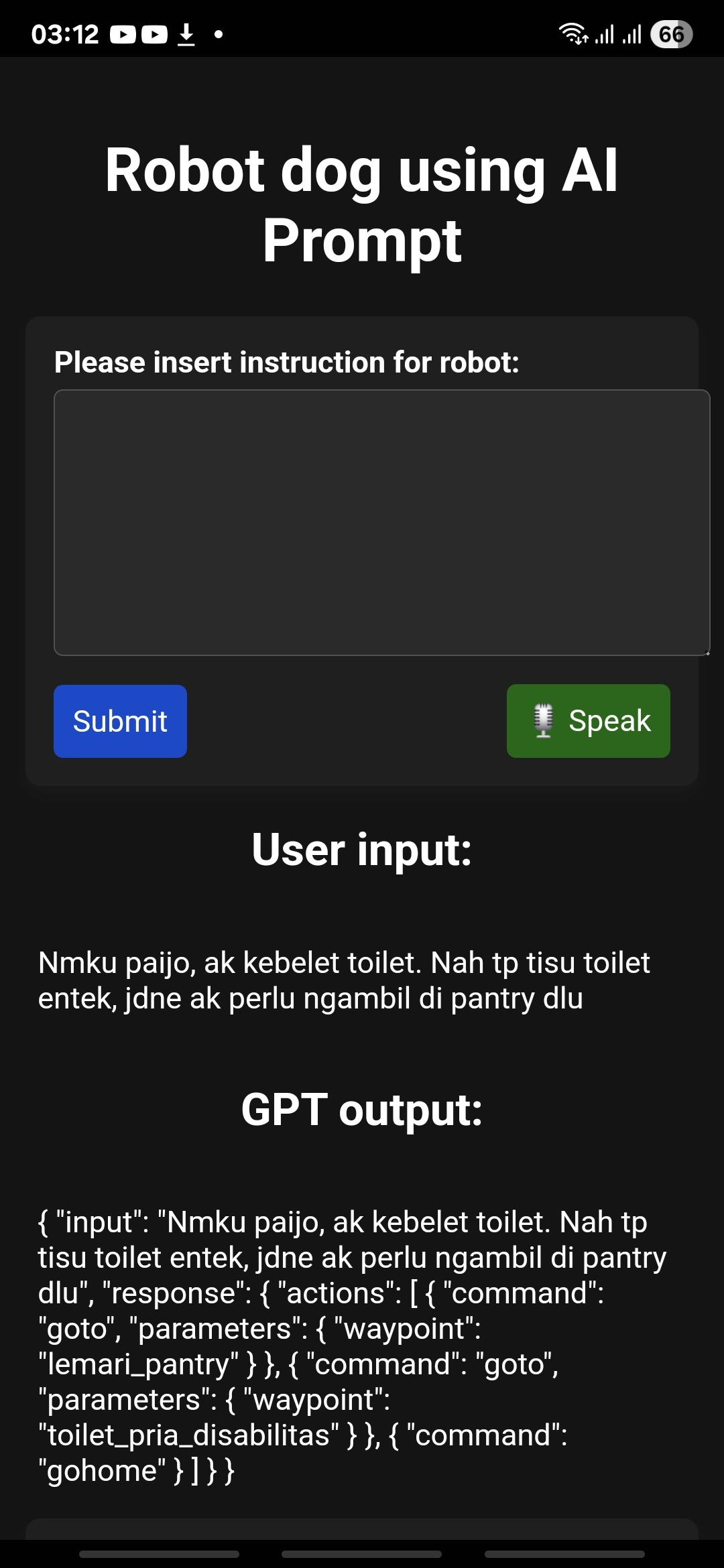}
\caption{Website interface for natural language input }
\label{fig:websiteinterface}
\end{figure}

The user interface is delivered through a responsive web application hosted on a Flask server running on the robot’s development host. This web interface allows users to input instructions in natural language and trigger the generation and execution of the movement plan. Fig. \ref{fig:websiteinterface} describes the web user interface in a mobile web browser layout.

Upon receiving input, the system calls a cloud-hosted LLM API to process the prompt and generate the corresponding JSON sequence. The resulting plan is parsed and relayed to the robot's motion planner via ROS topics, completing the cycle from user intention to robotic action.

\section{Experimental Results and Evaluation}
\subsection{Experimental Setup}
The proposed LLM-based robot control system was implemented and evaluated on the DeepRobotics Jueying Lite 3 quadruped-legged robot. The robot is equipped with two primary onboard computing modules: the motion host, responsible for direct communication with the motion actuators and sensors (LiDAR, IMU, odometry); and the perception host, which handles computationally intensive tasks such as localization, sensor fusion, and path planning. The perception host utilizes an NVIDIA Jetson Xavier NX to ensure real-time performance in indoor navigation tasks.

A third computing unit, the development host, is integrated into the system to manage interaction with the Large Language Model (LLM) and to host the web-based control interface via a Flask server. This server provides a user interface accessible through a web browser by any device on the same network.

The robot operates in a structured indoor environment: the 9th floor of Tower 2 at Institut Teknologi Sepuluh Nopember (ITS). This environment includes a variety of functional zones such as laboratories (TW901, TW903), hallways, pantry, restrooms, and elevators, all of which are mapped using ROS-compatible SLAM tools. Navigation and localization are handled using ROS and RViz, while the robot’s movement is managed via waypoints derived from the LLM-generated plans.

The experimental process consists of: (1) initializing robot sensors remotely via the perception host; (2) starting the ROS navigation stack and localizing the robot using RViz; (3) launching the LLM interface via a Flask server; and (4) sending natural language commands from the user through a browser to generate JSON-based movement plans.

\subsection{Test Scenarios}
The experiments are divided into four major categories based on the complexity and distance of movement tasks: single-room short navigation, multi-room short navigation, multi-room long navigation and cross-zone navigation. Each scenario evaluates the robot's ability to execute semantically complex commands derived from natural language, transformed into action plans by the LLM.
\begin{itemize}
\item \textbf{Single Room Short Navigation} : This scenario involved tasks within a single room inside Computer Engineering's 901 Laboratory. The robot achieved a 100\% success rate over 15 trials, with an average task completion time of 45.26 seconds.

\begin{example}{Single Room Navigation}{}
\textbf{Command}: \emph{Saya ingin mengambil barang di lemari lab, kemudian ingin menyoldernya.}
\vspace*{0.5cm}

Output : 
\vspace*{0.2cm}
\begin{lstlisting}[
    frame=single, 
    breaklines=true, 
    basicstyle=\ttfamily\small, 
    columns=fullflexible,
    captionpos=b,
    language=json]
{"response": {"actions": [{"command": "goto", "parameters": {"waypoint": "depan_lemari"}},{"command": "goto","parameters": {"waypoint": "depan_meja_solder"}}]}
\end{lstlisting}

\vspace{0.5cm}
\centering
\includegraphics[width=1\linewidth]{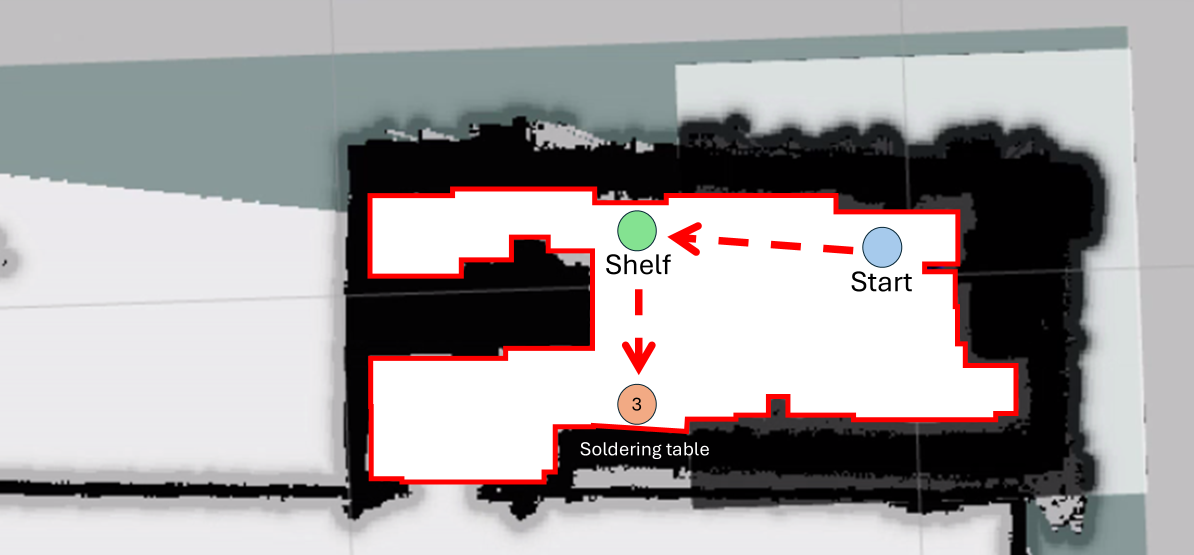}
\captionof{figure}{Path Generated}
\label{fig:singleroomnav}

\end{example}

\item \textbf{Multi-Room Short Navigation}: This involved simple transitions between two nearby areas, such as from 901 Laboratory to nearby 903 Laboratory and the Lifts located near Computer Engineering's Laboratory.  
This scenario has a 96\% success rate across 25 trials and an average time of 68.27 seconds

\begin{example}{Multi Room Short Distance}{}
\textbf{Command}: \emph{Saya ingin mengambil barang di lemari lab, kemudian juga mengambil barang di meja solder. Setelah itu saya ingin pergi ke lab TW903}
\vspace*{0.5cm}

Output : 
\vspace*{0.2cm}
% \begin{minted}[frame=single, breaklines]{json}
% { "response": { "actions": [ {"command": "goto", "parameters": {"waypoint": "depan_lemari"}}, {"command": "goto", "parameters": {"waypoint": "depan_meja_solder"}}, {"command": "goto", "parameters": {"waypoint": "depan_pintu_lab_903_luar"}} ] } }
% \end{minted}
\begin{lstlisting}[
    frame=single, 
    breaklines=true, 
    basicstyle=\ttfamily\small, 
    columns=fullflexible,
    captionpos=b,
    language=json]
{ "response": { "actions": [ {"command": "goto", "parameters": {"waypoint": "depan_lemari"}}, {"command": "goto", "parameters": {"waypoint": "depan_meja_solder"}}, {"command": "goto", "parameters": {"waypoint": "depan_pintu_lab_903_luar"}} ] } }
\end{lstlisting}

\vspace{0.5cm}
\centering
\includegraphics[width=0.92\linewidth]{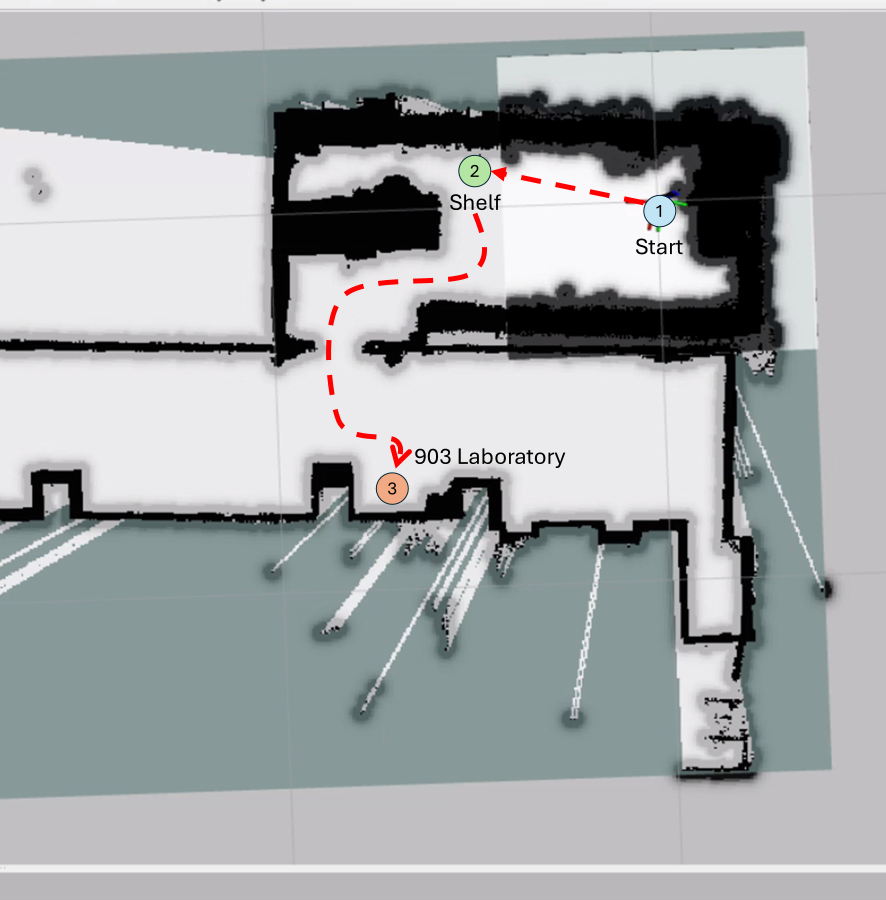}
\captionof{figure}{Path Generated}
\label{fig:multiroomshortnav}

\end{example}

% \item \textbf{Multi-Room Long Distance Navigation}: This required navigation between multi-room locations from the 901 Laboratory and farther parts of the 9th floor halls of Tower 2 ITS.
% Success rate was 90\% with an average duration of 89.71 seconds over 20 trials.

% \begin{example}{Multi Room Long Distance}
% \textbf{Command}: \emph{Saya ingin bertemu dengan teman saya di lab komputer, namun sebelum itu saya ingin mengambil keyboard yang ada di lemari lab dan saya solder. Setelah itu bertemu teman saya, saya akan turun ke lantai 2 dengan lift terdekat dari lab komputer}
% \vspace*{0.5cm}

% Output : 
% \vspace*{0.2cm}
% \begin{minted}[frame=single, breaklines]{json}
% {"response": { "actions": [ {"command": "goto", "parameters": {"waypoint": "depan_lemari"}}, {"command": "goto", "parameters": {"waypoint": "depan_meja_solder"}}, {"command": "goto", "parameters": {"waypoint": "lab_komputer_ti"}}, {"command": "goto", "parameters": {"waypoint": "lift_jauh"}} ] }}
% \end{minted}

% \vspace{0.5cm}
% \centering
% \includegraphics[width=1\linewidth]{multiroomlong.png}
% \captionof{figure}{Path Generated}
% \label{fig:multiroomlongnav}
% \end{example}

\item \textbf{Cross-Zone Navigation} : This scenario involved complex multi-zone on multiple area on the 9th floor of the Tower 2 ITS.  
Example: ``Saya ingin konsultasi ke lantai 2, tapi sebelumnya ambil hasil solderan dan pergi ke pantry serta toilet.''  
The robot succeeded in all 20 trials with an average task time of 130.98 seconds.

\begin{example}{Multi Room Long Distance} {}
\textbf{Command}: \emph{Ada acara halal bi halal di lantai 10. Namun sebelum itu, saya perlu mengambil sendok yang ada di lemari lab, kue di dalam pantry dan piring yang ada di lemari pantry. Saya ingin turun dengan lift terdekat dari pantry}
\vspace*{0.5cm}

Output : 
\vspace*{0.2cm}
\begin{lstlisting}[
    frame=single, 
    breaklines=true, 
    basicstyle=\ttfamily\small, 
    columns=fullflexible,
    captionpos=b,
    language=json]
{ "response": { "actions": [ { "command": "goto", "parameters": { "waypoint": "depan_lemari" } }, { "command": "goto", "parameters": { "waypoint": "ruang_pantry" } }, { "command": "goto", "parameters": { "waypoint": "lemari_pantry" } }, { "command": "goto", "parameters": { "waypoint": "lift_jauh" } } ] } }
\end{lstlisting}

\vspace{0.5cm}
\centering
\includegraphics[width=1\linewidth]{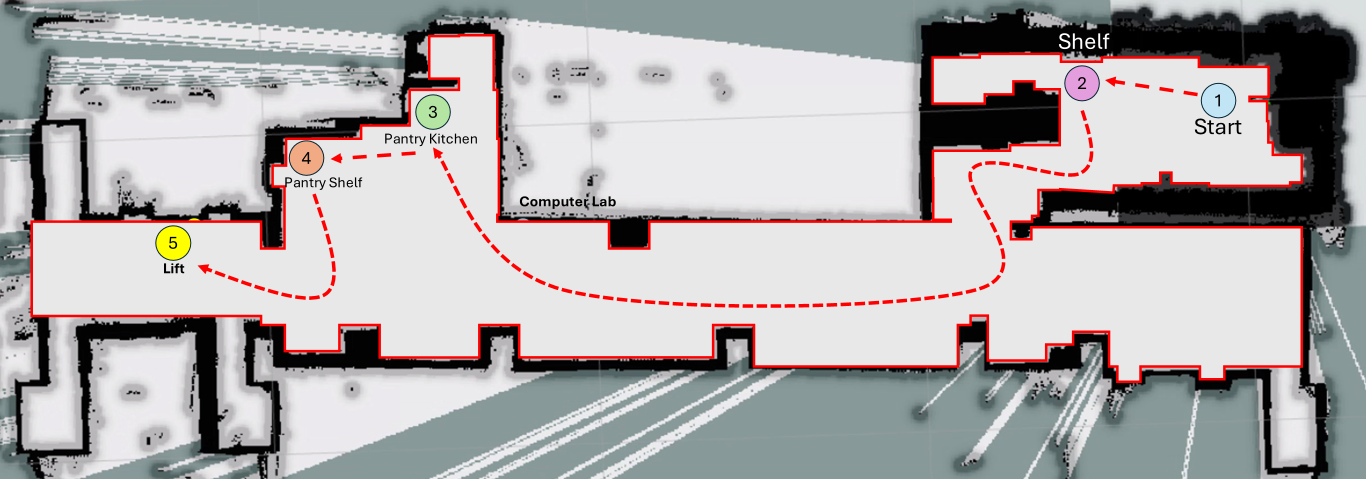}
\captionof{figure}{Path Generated}
\label{fig:CrosszoneNav}

\end{example}
\end{itemize}

\subsection{Performance Metrics}
To evaluate the performance of the proposed LLM-based control system for a quadruped-legged robot, we assessed several key metrics. These metrics aim to measure the system’s reliability in executing movement plans derived from natural language commands, as well as its efficiency in terms of execution time and success rate under varying levels of task complexity.
The main metrics evaluated include: (a). \textbf{Average Task Completion Time}: the duration from receiving a command until all planned actions are completed, and (b). \textbf{Success Rate}: the percentage of successful executions out of all attempts.

The test scenarios are categorized into three levels of complexity: single-room, multi-room, and cross-zone navigation. The summary of results is presented in Table \ref{tab:resultssummary}.

\begin{table}[htbp]
\caption{Summary of Experiment Results Across Scenario Categories}
\begin{center}
\resizebox{\columnwidth}{!}{%
\begin{tabular}{|l|c|c|c|}
\hline
\textbf{Scenario Category} & \textbf{Avg. Duration(s)} & \textbf{Success Rate(\%)} & \textbf{Total Attempts} \\
\hline
Short Dist. (Single-Room)  & 45.26  & 100 & 15 \\
Short Dist. (Multi-Room)   & 68.27  & 96  & 25 \\
Long Dist. (Multi-Room)    & 89.71  & 90  & 20 \\
Cross-Zone                 & 130.98 & 100 & 20 \\
\hline
\end{tabular}%
}
\label{tab:resultssummary}
\end{center}
\end{table}

\begin{figure}[htbp]
\centerline{\includegraphics[width=\columnwidth]{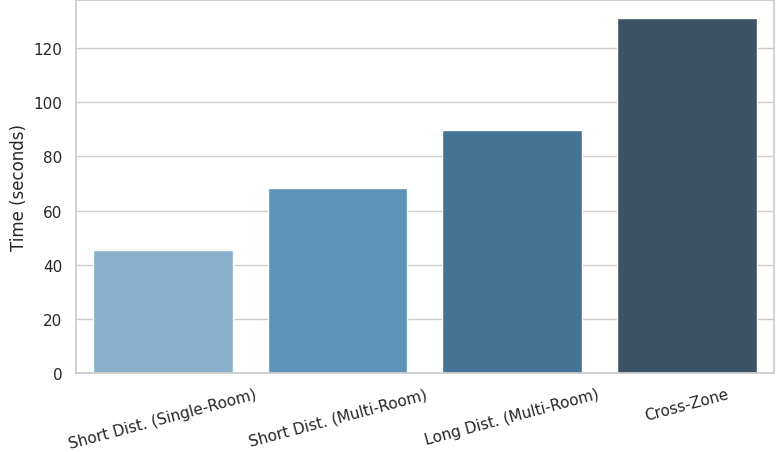}}
\caption{Average Task Completion Time by Scenario}
\label{fig:durationplot}
\end{figure}

\begin{figure}[htbp]
\centerline{\includegraphics[width=\columnwidth]{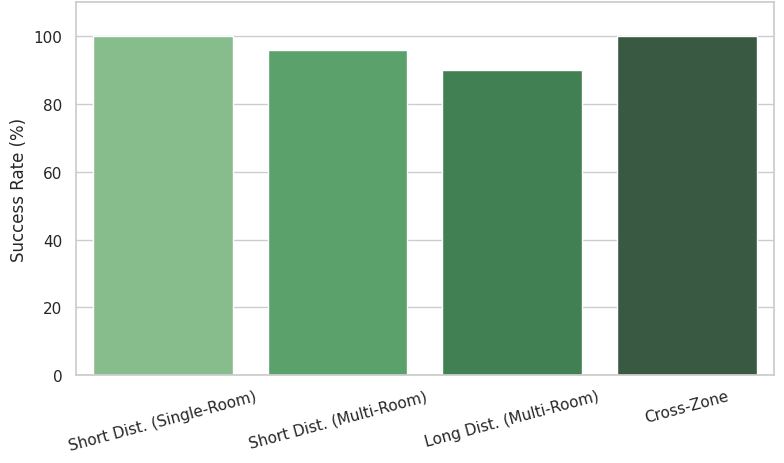}}
\caption{Success Rate by Scenario}
\label{fig:successplot}
\end{figure}

Figures~\ref{fig:durationplot} and~\ref{fig:successplot} illustrate the comparative performance in terms of task time and success rate. Based on  ~\ref{fig:durationplot} and~\ref{fig:successplot}, it clearly indicates a correlation between task complexity and execution duration. As expected, task time increases with the number of waypoints and the complexity of navigation paths. However, the consistently high success rates in both simple and complex tasks demonstrate the system’s robustness and reliability in interpreting and executing LLM-generated plans.

The failure in the multi-room scenario suggests opportunities for improvement in local navigation handling, such as path revalidation, map refinement, or better error recovery mechanisms during transition phases. Nevertheless, the consistent correctness of the generated JSON and absence of semantic errors indicate that the LLM performs reliably in transforming natural language into executable robot actions.

\section{conclusion}
We have presented a way to generate movement plans on Quadruped-Legged robots by integrating it with LLMs. Our methods enable robots to be easily controlled using any platform through human natural language, without prior skills or programming knowledge, and operate a robot naturally. This
method also allows an easier way to program a robot in a new environment, with requirements that include a map, global coordinates, and LLM prompt. Our methods have proven to be reliable for generating movement plans. For future works, we plan to integrate retrieval augmented generation(RAG) as a way for the robot to take context from all the previous prompts the user makes. We also plan to integrate Visual Language Model as a way for the robot to caption and know its surrounding environment to make adjustments based on visual input.

\section*{Acknowledgment}
This research was financially supported by the Final Project Assistance Grant (Bantuan Tugas Akhir Mahasiswa) funded by Institut Teknologi Sepuluh Nopember (ITS) under the 2025 ITS Internal Research Grant Scheme.


\begin{thebibliography}{00}

\bibitem{Zhou_Adaptive} K. Zhou, Y. Mu, H. Song, et al., Adaptive interactive navigation of quadruped robots using large language models, 2025. arXiv: 2503.22942 [cs.RO]. [Online]. Available: https://arxiv.org/abs/2503.22942.

\bibitem{Han_Lifelike} L. Han, Q. Zhu, J. Sheng, et al., “Lifelike agility and play in quadrupedal robots using reinforcement learning and generative pre-trained models,” Nature Machine Intelligence, vol. 6, pp. 787–798, 7 Jul. 2024, ISSN: 2522-5839. DOI: 10 . 1038 / s42256 - 024 - 00861 - 3. [Online]. Available: https://doi.org/10.1038/s42256-024-00861-3.

\bibitem{Mastalli_Motion} C. Mastalli, I. Havoutis, M. Focchi, D. G. Caldwell,and C. Semini, “Motion planning for quadrupedal locomotion: Coupled planning, terrain mapping, and whole-body control,” IEEE Transactions on Robotics, vol. 36, pp. 1635–1648, 6 Dec. 2020, ISSN: 1552-3098, 1941-0468. DOI: 10 . 1109 / tro . 2020 . 3003464. [Online]. Available:http://dx.doi.org/10.1109/tro.2020.3003464.

\bibitem{Nwankwo_conversation} L. Nwankwo and E. Rueckert, “The conversation is the command: Interacting with real-world autonomous robot through natural language,” arXiv (Cornell Uni-versity), Jan. 2024. DOI: 10.48550/arxiv.2401.11838. [Online]. Available: http://arxiv.org/abs/2401.11838.

\bibitem{Wilde_Improving} N. Wilde, A. Blidaru, S. L. Smith, and D. Kuli´c, “Improving user specifications for robot behavior through active preference learning: Framework and evaluation,” The International Journal of Robotics Research, vol. 39, pp. 651–667, 6 Mar. 2020, ISSN: 0278-3649, 1741-3176. DOI: 10 . 1177 / 0278364920910802. [Online]. Available: https://doi.org/10.1177/0278364920910802.

\bibitem{Cai_Low-code} Y. Cai, S. Mao, W. Wu, et al., “Low-code llm: Graphical user interface over large language models,” arXiv (Cornell University), Apr. 2023. DOI: 10.48550/arxiv.2304.08103. [Online]. Available: http://arxiv.org/abs/2304.08103.

\bibitem{Ahn_Do} M. Ahn, A. Brohan, N. Brown, et al., Do as i can, not as i say: Grounding language in robotic affordances, 2022. arXiv: 2204 . 01691 [cs.RO]. [Online]. Available: https://arxiv.org/abs/2204.01691.

\bibitem{Ginting_comply} M. F. Ginting, D.-K. Kim, S.-K. Kim, et al., Say comply: Grounding field robotic tasks in operational compliance through retrieval-based language models, 2024. arXiv: 2411 . 11323 [cs.RO]. [Online]. Available: https://arxiv.org/abs/2411.11323.

\bibitem{Balaji_Language} A. Balaji, S. Pradhan, and D. Berenson, Language-guided object search in agricultural environments,2025. arXiv: 2503 . 01068 [cs.RO]. [Online]. Available: https://arxiv.org/abs/2503.01068.

\bibitem{Anwar_Remembr} A. Anwar, J. Welsh, J. Biswas, S. Pouya, and Y. Chang, "Remembr: Building and reasoning over long-horizon spatio-temporal memory for robot navigation", 2024. arXiv: 2409 . 13682 [cs.RO]. [Online]. Available:https://arxiv.org/abs/2409.13682.

\bibitem{Waga_survey} A. Waga, S. Benhlima, A. Bekri, J. Abdouni, and F. Z.Saber, “A survey on autonomous navigation for mobile robots: From traditional techniques to deep learning and large language models,” Journal of King Saud University - Computer and Information Sciences, vol. 37, 7 Aug. 2025, ISSN: 1319-1578, 2213-1248. DOI: 10.1007/s44443-025-00216-x. [Online]. Available: https://doi.
org/10.1007/s44443-025-00216-x.

\bibitem{Lin_Advances} J. Lin, H. Gao, R. Xu, C. Wang, L. Guo, and S. Xu, “Advances in embodied navigation using large language models: A survey,” arXiv (Cornell University), Nov. 2023. DOI: 10 . 48550 / arxiv . 2311 . 00530. [Online]. Available: http://arxiv.org/abs/2311.00530.

\bibitem{Muhtadin_Autonomous} Muhtadin, R. M. Zanuar, I. K. E. Purnama, and M. H.Purnomo, “Autonomous navigation and obstacle avoidance for service robot,” in 2019 International Conference on Computer Engineering, Network, and Intelligent Multimedia (CENIM), 2019, pp. 1–8. DOI: 10 .
1109/CENIM48368.2019.8973360.

\bibitem{Muhtadin_Implementation} I. K. E. Purnama, M. A. Pradana, and Muhtadin,“Implementation of object following method on robot service,” in 2018 International Conference on Computer Engineering, Network and Intelligent Multimedia (CENIM), 2018, pp. 172–175. DOI: 10.1109/CENIM.
2018.8710819.

\bibitem{Muhtadin_Robot} Muhtadin, Billy, E. M. Yuniarno, et al., “Robot service for elderly to find misplaced items: A resource efficient implementation on low-computational device,”in 2020 IEEE International Conference on Industry 4.0, Artificial Intelligence, and Communications Technology (IAICT), 2020, pp. 28–34. DOI: 10.1109/IAICT50021.2020.9172030.


\end{thebibliography}
\end{document}